\documentclass{article}

\PassOptionsToPackage{numbers, compress}{natbib}

    \usepackage[preprint]{neurips_2021}

\usepackage[utf8]{inputenc} 
\usepackage[T1]{fontenc}    
\usepackage{hyperref}       
\usepackage{url}            
\usepackage{booktabs}       
\usepackage{amsfonts}       
\usepackage{nicefrac}       
\usepackage{microtype}      
\usepackage{xcolor}         

\makeatletter
\newcommand{\leqnomode}{\tagsleft@true\let\veqno\@@leqno}
\newcommand{\reqnomode}{\tagsleft@false\let\veqno\@@eqno}
\makeatother

\usepackage{amsmath}

\DeclareMathOperator*{\minimize}{\text{minimize}}

\DeclareMathOperator*{\st}{\text{subject to}}
\DeclareMathAlphabet\mathbfcal{OMS}{cmsy}{b}{n}

\newcommand{\bz}{\mathbf z}

\newcommand{\btheta}{\boldsymbol{\theta}}

\usepackage[pdftex]{graphicx}

\title{Human-Centered AI for Data Science: \\ A Systematic Approach}

\author{%
  Dakuo Wang \thanks{Corresponding author.} \\
  IBM Research\\
  \texttt{dakuo.wang@ibm.com} \\
   \And
   Xiaojuan Ma \\
    Hong Kong University of \\ Science and Technology\\
   \texttt{mxj@cse.ust.hk} \\
   \And
   April Yi Wang \\
   University of Michigan \\
   \texttt{aprilww@umich.edu} \\
}

\begin{document}

\maketitle
\vspace{-10pt}
\begin{abstract}
  Human-Centered AI (HCAI) refers to the research effort that aims to design and implement AI techniques to support various human tasks, while taking human needs into consideration and preserving human control. 
  In this short position paper, we illustrate how we approach HCAI using a series of research projects around Data Science (DS) works as a case study. 
  The AI techniques built for supporting DS works are collectively referred to as AutoML systems, and their goals are to automate some parts of the DS workflow.
  We illustrate a \textbf{three-step systematical research approach }(i.e., explore, build, and integrate) and \textbf{four practical ways of implementation} for HCAI systems. 
  We argue that our work is a cornerstone towards the ultimate future of Human-AI Collaboration for DS and beyond, where AI and humans can take complementary and indispensable roles to achieve a better outcome and experience.  
\end{abstract}

\section{Introduction and Background}
Neural network-based AI techniques have been a popular research topic in recent years, but more and more researchers realize that there is a huge gap between an AI algorithm research paper and a human-usable AI system.
Many researchers across different fields (e.g., Human Computer Interaction, AI, etc.) have started to form a new Human-Centered AI (HCAI) research agenda. The signature of HCAI is to design and build AI algorithms and systems around real human workflow and user needs, so that the AI systems can indeed augment human capability in a controllable and trustable manner for different tasks~\cite{kogan2020mapping,wang2021autods,drozdal2020trust}.

\begin{figure}[t]
    \centering
    \includegraphics[width=0.5\linewidth]{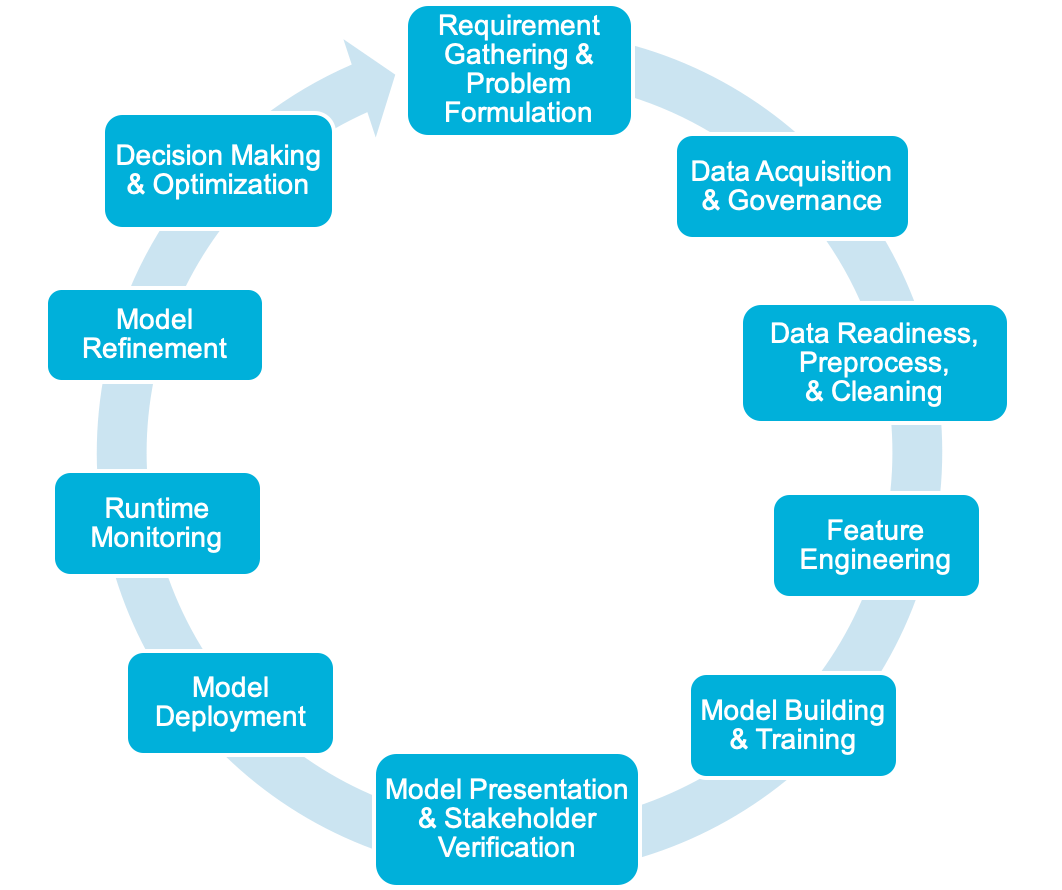}
      \caption{Ten Typical Tasks in a DS Lifecycle~\cite{wang2021autods}}
      \label{fig:lifecycle}
\end{figure}

We use the data science (DS) context as a case study to demonstrate possible paths to HCAI.
Data science refers to the practice of using machine learning (ML) modeling techniques to generate domain-specific insights from data and then making better decisions with these data-driven insights.
It is known that DS work has the following attributes: 1) it is an iterative multi-step workflow~\cite{wang2021autods}, as shown in Fig. \ref{fig:lifecycle}; 2) it is a collaborative task that involves multiple stakeholders~\cite{zhang2020data}; 3) it is an interdisciplinary work that team members with different backgrounds work together~\cite{Muller:2019:DSW:3290605.3300356,mao2019}; and 4) while the ML modeling part of the work is tedious, the non-modeling tasks (e.g., data preparation or communication) actually take up most of the time (about 80\%)~\cite{8020dilemma}. 

Using AI automation to speed up and scale up the various works in the DS lifecyle has been popular in the last few years~\cite{wang2021much,xin2021whither}. 
Researchers have invented various AI optimization techniques to automatically select a model algorithm~\cite{he2021automl}, generate new features~\cite{galhotra2019automated}, try out candidate values of hyperparameters ~\cite{liu2020admm}, and facilitate some other stages of the DS project lifecycle~\cite{zhang2020data}. 
People refer to this group of techniques as automated machine learning (AutoML).


We focus on this AutoML type of AI techniques and illustrate how we research and build HCAI systems to support various stakeholders and tasks in the DS lifecycle. 

\vspace{-5pt}

\section{Explore, Build, and Integrate Human-Centered AI for Data Science}

In this position paper, we demonstrate a three-step systematical research approach that we used to approach the HCAI research in DS: we first conduct exploration through user studies to understand human teams' work practices to serve as a baseline for potential AI intervention; then, we design and build AI applications and algorithms; lastly, we integrate and evaluate AI systems within humans' real-world workflows. 
\subsection{Explore Human Workflow and User Needs}
In order to build a HCAI system that can actually augment DS workers' capability, we first need to gather knowledge on how humans work while they do not have help from AI. 
Prior research \cite{aragon2016developing, kogan2020mapping, muller2019human,wang2021much} have found DS often consists of multiple stages in a project lifecycle: from gathering requirements and datasets, to
deploying a model, and to making decisions; and users do not want AI to automate the entire lifecycle. Researchers have also found there are diverse personas in the DS/ML team \cite{zhang2020data} and these personas must collaborate on various tasks in the lifecycle. For example, stakeholders set requirements with data scientists, data engineers support data cleaning and model building, and later, stakeholders verify the model and domain experts use model inferences in decision making, and so on~\cite{Muller:2019:DSW:3290605.3300356}. 

We summarize the above insights into the following four design requirements for HCAI systems in the context of DS: \\
1) HCAI systems should not aim to automate the entire DS lifecycle, but should specify which DS task(s) it aims to support and with what levels of automation. \\
2) HCAI systems should specify which user personna it aims to support for the target task(s). \\
3) HCAI systems should consider the interdisciplinary background of users. \\
4) The modeling-related tasks are easier for AI, whereas the human-knowledge-driven tasks are harder for AI to support.

\subsection{Build Human-Centered AI Systems}
These prior literature findings and our own formative studies can support the HCAI systems design decisions. We decompose the AI lifecycle into different tasks and each of our following HCAI systems targets one particular task, with which we illustrate the four practical ways of HCAI implementation. 

\begin{figure}[t]
    \centering
    \includegraphics[width=0.6\linewidth]{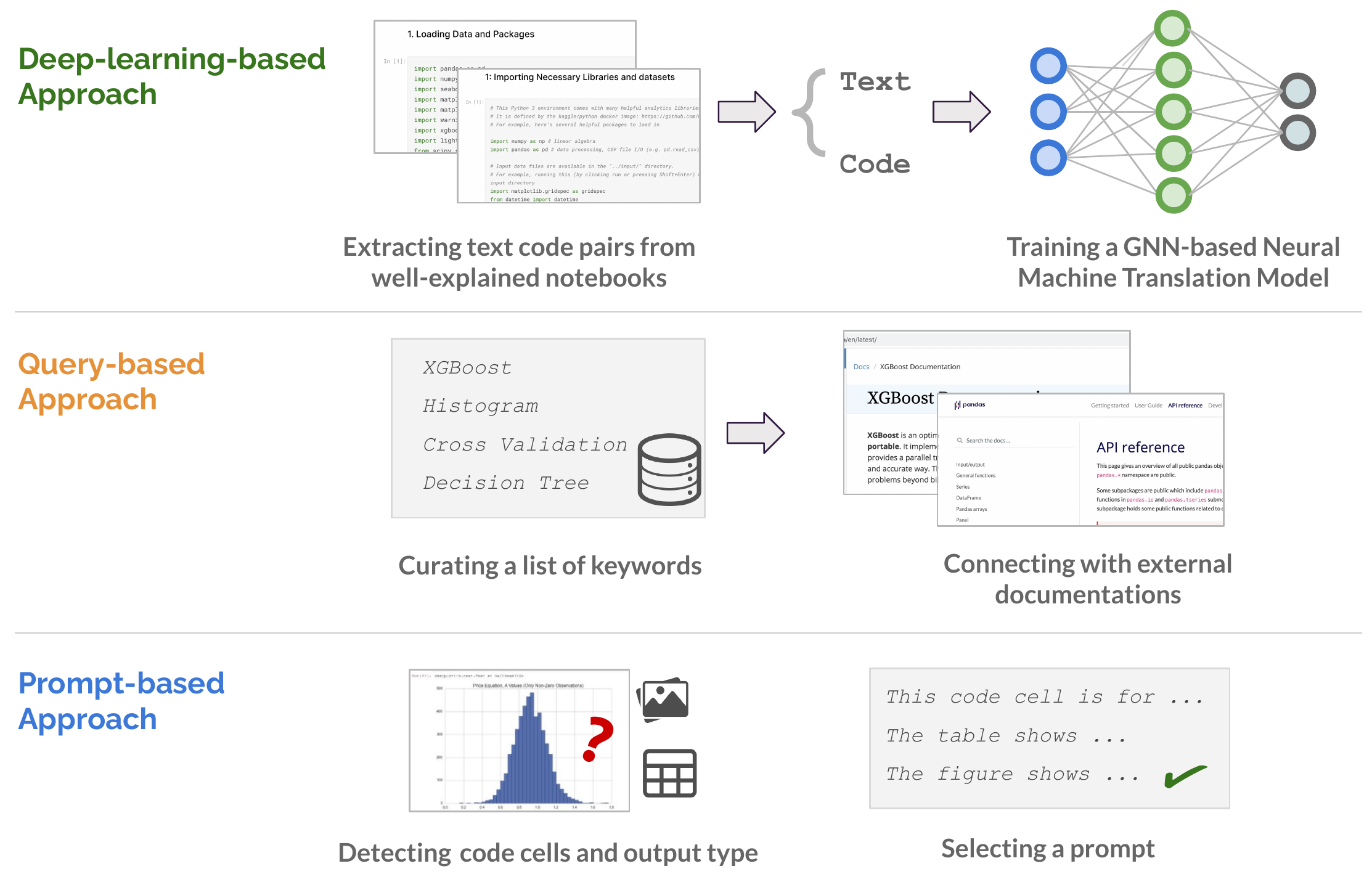}
    \caption{A hybrid system architecture with both rule-based and AI-based approaches for DS code documentation generation.}
    \label{fig:docgen}
\end{figure}

\paragraph{A hybrid system with human-crafted rules and AI.}
The first exemplar HCAI system aims to automate the DS code documentation task, as a good documentation in code can help DS workers to share their and reuse other's work~\cite{rule2018exploration, zhang2020data}. 
Because DS workers often neglect to do so~\cite{rule2018exploration}, we built an AI system to support them.
After examining human good practices of code documentation, we found that people document their DS code for different purposes, and some of those documentation types (e.g., interpreting results) are too challenging for state-of-art AI algorithmes to generate. 
Thus, we designed a hybrid system architecture (Fig.~\ref{fig:docgen}) -- when it reaches the AI's limit, the system can let users know AI fails and prompt them to step in to manually write the documentation. 

\begin{figure}
\centering
  \includegraphics[width=0.7\columnwidth]{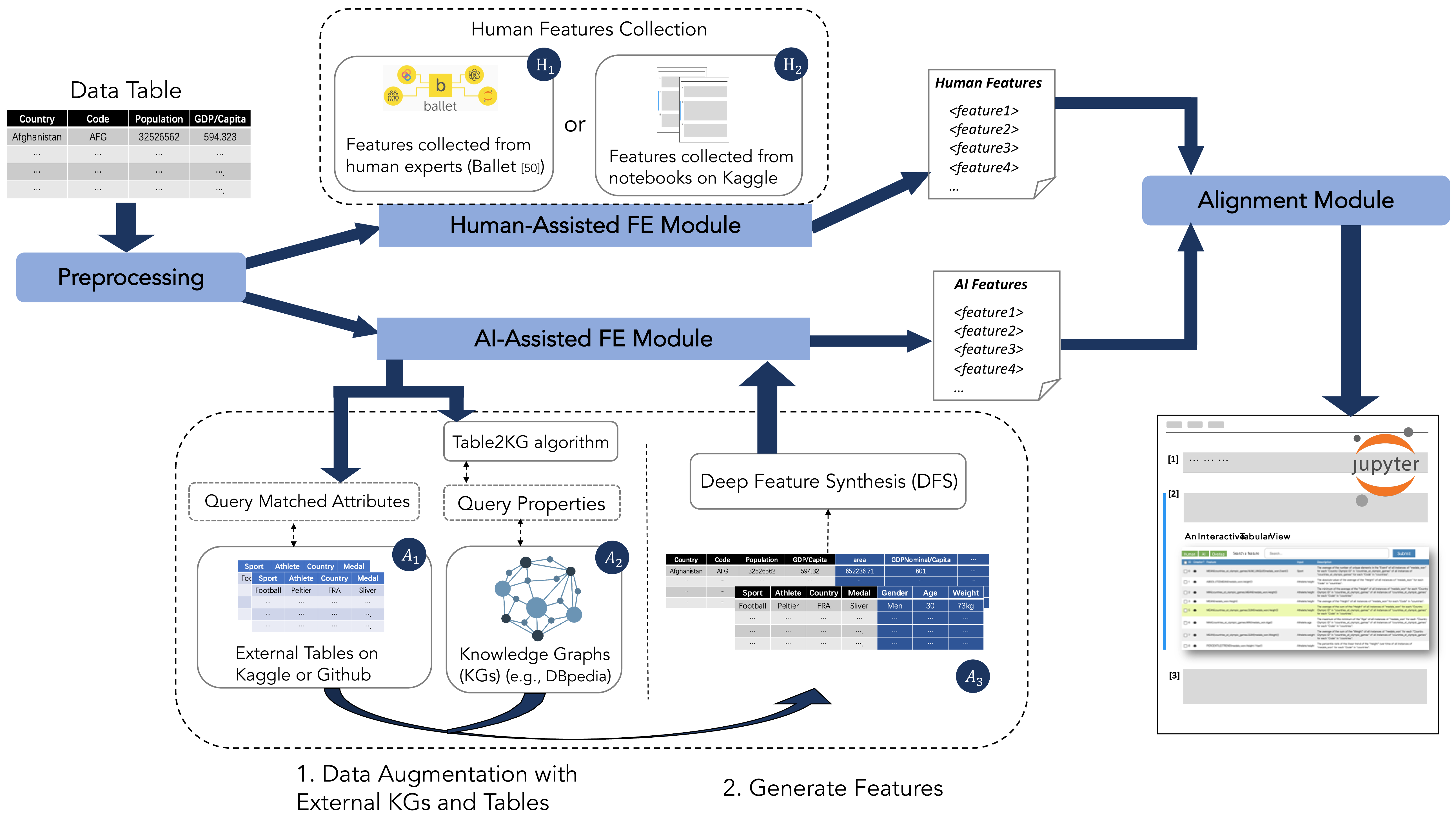}
  \caption{An overview of the system architecture that has human crowdworker-in-the-loop as a human-assisted FE module in parallel to the AI-assisted FE module.} 
  \label{fig:fe}
\end{figure}

\paragraph{Human-in-the-loop AI system.}
The second examplar HCAI system adopts a human-in-the-loop design approach to support the feature engineering (FE) task in DS lifecycle.
FE refers to the creation and selection of features to improve the final model's performance. For example, the original dataset may only has a person's body weight and body height information, and FE can create one new feature body-mass-index (BMI) using the two existing features.
Despite many AI researchers are actively working on the automated solutions for FE, we argue that FE is a highly human-knowledge-driven task and thus we decided to build a human-in-the-loop architecture to leverage both AI and human crowd workers to support users' FE task (Fig.~\ref{fig:fe}).

\paragraph{Human needs as AI model's optimization constraints.}
The model training and model selection task in the AI lifecycle has been the primary focus of AutoML research, and hence AI's performance is reasonable for this task\footnote{Let $f\left(\bz, \btheta; \mathcal{A} \right)$ denote some notion loss of a ML pipeline corresponding to the algorithm choices as per $\bz$ with the hyper-parameters $\btheta$ on a learning task with data $\mathcal{A}$}(solving equation~\ref{eq: prob0}). However, we found that DS workers in practice also need to consider business requirements in addition to the model's prediction accuracy. For example, a loan approval AI system needs to make sure the model does not discriminates different group of applicants (e.g., performing very well on male applicants but very poorly on female applicants), and it needs to calculate the results within a very short amount of time (less than 10 ms). 
{\small\begin{equation}\label{eq: prob0}
\minimize_{\bz, \btheta} f(\bz,  \btheta; \mathcal{A}) 
\st
\left\{
\begin{array}{l}
\bz_i \in \{0,1 \}^{K_i}, \mathbf 1^{\top} \bz_i = 1, \forall i \in [N], \\
\btheta_{ij}^c \in \mathcal{C}_{ij}, \btheta_{ij}^d \in \mathcal{D}_{ij} 
\forall i \in [ N ], j \in [ K_i ].
\end{array}
\right.
\end{equation}}%

We incorporate such requirements by extending our AutoML equation \eqref{eq: prob0} to include $M$ {\em black-box constraints such as fairness and prediction time}: 
%
{\small\begin{equation}
    \label{eq: prob0-csts}
    g_i\left(\bz, \btheta; \mathcal{A} \right) \leq \epsilon_i, i = 1, \ldots, M.
 \end{equation}}%
and incorporate these constraints (equation~\ref{eq: prob0-csts}) into the black-box objective with some penalty function $p(\cdot)$, where the new objective becomes $f + \sum_i p(g_i, \epsilon_i)$ or $f \cdot \prod_i p(g_i, \epsilon_i)$, and use a Bayesian optimization-based approach to solve it.

\begin{figure}[h]
  \centering
  \includegraphics[width=0.9\linewidth]{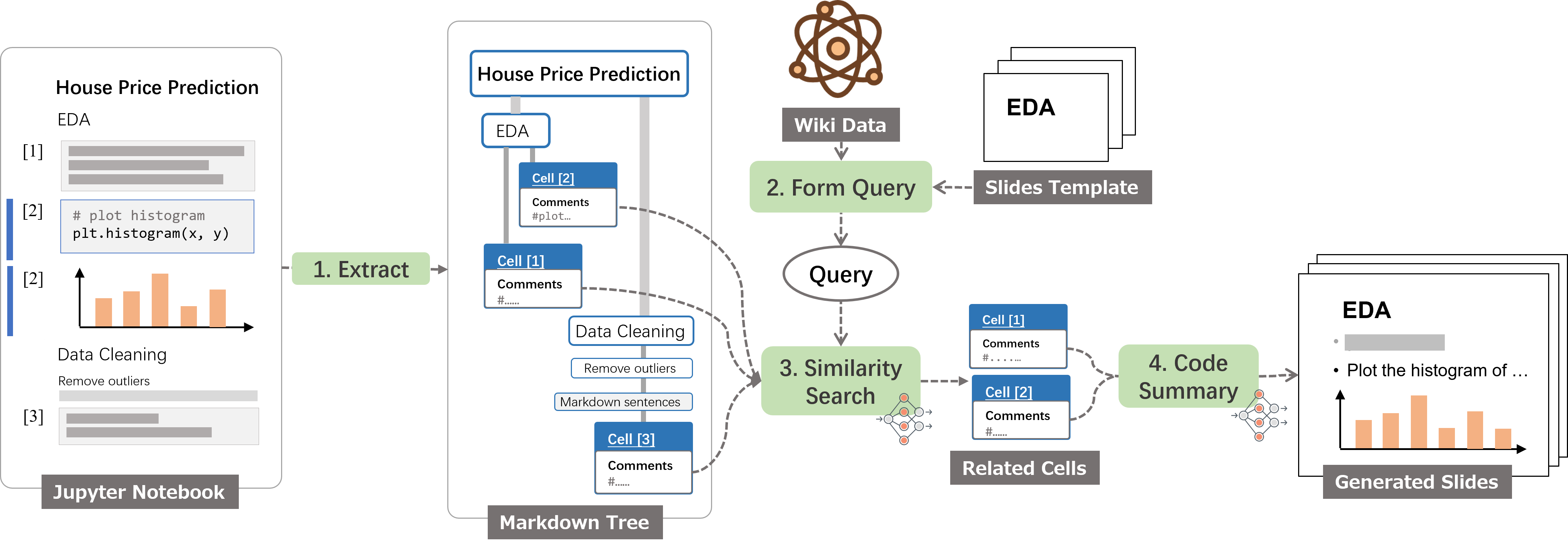}
  \caption{An system architecture that incorporates human knowledge representation and neural network for generating presentation slides.}
  \label{fig:nb2slide}
\end{figure}

\paragraph{Neural-symbolic modeling.}
Neurosymbolic models are composed of both symbolic representations such as rules and conditionals and continuous neural network components. The symbolic part represents human knowledge thus makes the model interpretable and
generalizable, while the neural part handles the diversity.
As an example, we build an HCAI system that focuses on the DS presentation task in the AI lifecycle --- it can automatically generate presentation slides from jupyter notebook codes. 
The architecture (Fig~\ref{fig:nb2slide}) leverage human knowledge representation in the symbolic slides template and wiki database, with which it calculates a similarity score for the notebook code and the user-expected presentation structure. Then it follows a neural-network based seq2seq model for generating code summary.

\vspace{-5pt}

\subsection{Integrate HCAI Systems into Human Workflow}
AI researchers are used to use automated evaluation metrics such as accuracy and F1 scores to evaluate their model and system performance. 
Fortunately, more and more researcher become aware of the unique power and necessity of human evaluation on top of the automated evaluation~\cite{ribeiro2020beyond}. 
We argue that all HCAI system works should have a follow-up user evaluation, at least in the lab but better in a real-world adoption.
That is why we have evaluated all the four exemplar systems with DS workers.
Some parts of our work (e.g., the black-box optimization constraints) have been implemented into a product that thousands of users are using in their daily DS works.
We would like to remind the readers that the user perception and societal impact of HCAI systems are also important research topics.
An example prior work is that researchers conducted an interview study and reported that DS workers believe that automation can assist them, but they do not worry their job security at all, as these AI systems can never (and should not be designed to) replace their jobs~\cite{wang2019human}. 


\section{Conclusion}
In this position paper, we join the interdisciplinary research effort of human centered AI research. 
Using data science as a case study context, we demonstrate a systematical HCAI research approach with three steps: explore, build, and integrate. 
In particular, based on our practice, we summarized four ways to build HCAI systems. 
We hope through this work, we can inspire readers to explore more opportunities of the HCAI line of research.
Together, we can research and build HCAI systems towards the ultimate human-AI collaboration future of work.



\medskip
\bibliographystyle{abbrv}
{
\small
\bibliography{ref}

\begin{thebibliography}{10}

\bibitem{aragon2016developing}
C.~Aragon, C.~Hutto, A.~Echenique, B.~Fiore-Gartland, Y.~Huang, J.~Kim,
  G.~Neff, W.~Xing, and J.~Bayer.
\newblock Developing a research agenda for human-centered data science.
\newblock In {\em Proceedings of the 19th ACM Conference on Computer Supported
  Cooperative Work and Social Computing Companion}, pages 529--535. ACM, 2016.

\bibitem{drozdal2020trust}
J.~Drozdal, J.~Weisz, D.~Wang, G.~Dass, B.~Yao, C.~Zhao, M.~Muller, L.~Ju, and
  H.~Su.
\newblock Trust in automl: Exploring information needs for establishing trust
  in automated machine learning systems.
\newblock In {\em Proceedings of the 25th International Conference on
  Intelligent User Interfaces}, pages 297--307, 2020.

\bibitem{galhotra2019automated}
S.~Galhotra, U.~Khurana, O.~Hassanzadeh, K.~Srinivas, H.~Samulowitz, and M.~Qi.
\newblock Automated feature enhancement for predictive modeling using external
  knowledge.
\newblock In {\em 2019 International Conference on Data Mining Workshops
  (ICDMW)}, pages 1094--1097. IEEE, 2019.

\bibitem{he2021automl}
X.~He, K.~Zhao, and X.~Chu.
\newblock Automl: A survey of the state-of-the-art.
\newblock {\em Knowledge-Based Systems}, 212:106622, 2021.

\bibitem{kogan2020mapping}
M.~Kogan, A.~Halfaker, S.~Guha, C.~Aragon, M.~Muller, and S.~Geiger.
\newblock Mapping out human-centered data science: Methods, approaches, and
  best practices.
\newblock In {\em Companion of the 2020 ACM International Conference on
  Supporting Group Work}, pages 151--156, 2020.

\bibitem{liu2020admm}
S.~Liu, P.~Ram, D.~Vijaykeerthy, D.~Bouneffouf, G.~Bramble, H.~Samulowitz,
  D.~Wang, A.~Conn, and A.~Gray.
\newblock An admm based framework for automl pipeline configuration.
\newblock In {\em Proceedings of the AAAI Conference on Artificial
  Intelligence}, volume~34, pages 4892--4899, 2020.

\bibitem{mao2019}
Y.~Mao, D.~Wang, M.~Muller, K.~Varshney, I.~Baldini, C.~Dugan, and
  A.~Mojsilovic.
\newblock How data scientists work together with domain experts in scientific
  collaborations.
\newblock In {\em Proceedings of the 2020 ACM conference on GROUP}. ACM, 2020.

\bibitem{muller2019human}
M.~Muller, M.~Feinberg, T.~George, S.~J. Jackson, B.~E. John, M.~B. Kery, and
  S.~Passi.
\newblock Human-centered study of data science work practices.
\newblock In {\em Extended Abstracts of the 2019 CHI Conference on Human
  Factors in Computing Systems}, page W15. ACM, 2019.

\bibitem{Muller:2019:DSW:3290605.3300356}
M.~Muller, I.~Lange, D.~Wang, D.~Piorkowski, J.~Tsay, Q.~V. Liao, C.~Dugan, and
  T.~Erickson.
\newblock How data science workers work with data: Discovery, capture,
  curation, design, creation.
\newblock In {\em Proceedings of the 2019 CHI Conference on Human Factors in
  Computing Systems}, CHI '19, pages 126:1--126:15, New York, NY, USA, 2019.
  ACM.

\bibitem{ribeiro2020beyond}
M.~T. Ribeiro, T.~Wu, C.~Guestrin, and S.~Singh.
\newblock Beyond accuracy: Behavioral testing of nlp models with checklist.
\newblock {\em arXiv preprint arXiv:2005.04118}, 2020.

\bibitem{8020dilemma}
A.~Ruiz and A.~Ruiz.
\newblock The 80/20 data science dilemma, Sep 2017.

\bibitem{rule2018exploration}
A.~Rule, A.~Tabard, and J.~D. Hollan.
\newblock Exploration and explanation in computational notebooks.
\newblock In {\em Proceedings of the 2018 CHI Conference on Human Factors in
  Computing Systems}, pages 1--12, 2018.

\bibitem{wang2021autods}
D.~Wang, J.~Andres, J.~Weisz, E.~Oduor, and C.~Dugan.
\newblock Autods: Towards human-centered automation of data science.
\newblock {\em Proceeding of CHI'21}, 2021.

\bibitem{wang2021much}
D.~Wang, Q.~V. Liao, Y.~Zhang, U.~Khurana, H.~Samulowitz, S.~Park, M.~Muller,
  and L.~Amini.
\newblock How much automation does a data scientist want?
\newblock {\em arXiv preprint arXiv:2101.03970}, 2021.

\bibitem{wang2019human}
D.~Wang, J.~D. Weisz, M.~Muller, P.~Ram, W.~Geyer, C.~Dugan, Y.~Tausczik,
  H.~Samulowitz, and A.~Gray.
\newblock Human-ai collaboration in data science: Exploring data scientists'
  perceptions of automated ai.
\newblock {\em Proceedings of the ACM on Human-Computer Interaction},
  3(CSCW):1--24, 2019.

\bibitem{xin2021whither}
D.~Xin, E.~Y. Wu, D.~J.-L. Lee, N.~Salehi, and A.~Parameswaran.
\newblock Whither automl? understanding the role of automation in machine
  learning workflows.
\newblock {\em arXiv preprint arXiv:2101.04834}, 2021.

\bibitem{zhang2020data}
A.~X. Zhang, M.~Muller, and D.~Wang.
\newblock How do data science workers collaborate? roles, workflows, and tools.
\newblock {\em arXiv preprint arXiv:2001.06684}, 2020.

\end{thebibliography}
}




\end{document}